\documentclass[twocolumn,twoside,fancyhea]{aeb02}
\usepackage[dvips]{epsfig}
\usepackage{fancyheadings}

\def\BibTeX{{\rm B\kern-.05em{\sc i\kern-.025em b}\kern-.08em
    T\kern-.1667em\lower.7ex\hbox{E}\kern-.125emX}}

\begin{document}

% \pagestyle{fancyplain}
% \thispagestyle{fancyplain}
% \headrulewidth 0pt
% \lhead[Primer congreso iberoamericano de algoritmos evolutivos y bioinspirados, AEB\'{}02, M\'erida, Feb.2002]{}
% \rhead[]{P. A. Castillo et al.: Lamarckian Evolution and the Baldwin Effect ...}
% \lfoot[]{}
% \cfoot[]{}
% \rfoot[]{}

\title{Lamarckian Evolution and the Baldwin Effect in Evolutionary Neural Networks}

\author {\normalsize{P.A. Castillo$^1$, M.G. Arenas$^1$, J.G. Castellano$^1$, J.J. Merelo$^1$, A. Prieto$^1$, V. Rivas$^2$ and G. Romero$^1$}
\thanks{Grupo Geneura
    $^1$Departamento de Arquitectura y Tecnolog\'{\i}a de Computadores. 
    Universidad de Granada. 
    Campus de Fuentenueva 
    18071 Granada (Spain).
    $^2$Departamento de Inform\'atica.
    Universidad de Ja\'en.
    E.P.S., Avda. Madrid, 35
    23071 Ja\'{e}n (Spain).
    e-mail: {\tt todos@geneura.ugr.es}
    URL: {\tt http://www.geneura.org} }}

%%%%%%%%%%%%%%%%%%%%%%%%%%%%%%%%%%%%%%%%%%%%%%%%%%%%%%%%%%%%%%%%%%%%%%%%%%%%%%%
\maketitle

%\markboth
%{P. A. Castillo et al.: Lamarckian Evolution and the Baldwin Effect ...}
%{Primer congreso iberoamericano de algoritmos evolutivos y bioinspirados, AEB\'{}02, M\'erida, Feb.2002}

%%%%%%%%%%%%%%%%%%%%%%%%%%%%%%%%%% Abstract %%%%%%%%%%%%%%%%%%%%%%%%%%%%%%%%%%%
\begin{abstract}
Hybrid neuro-evolutionary algorithms may be inspired on Darwinian or Lamarckian evolution. In the case of Darwinian evolution, the Baldwin effect, that is, the progressive incorporation of learned characteristics to the genotypes, can be observed and leveraged to improve the search.

The purpose of this paper is to carry out an experimental study into how learning can improve G-Prop genetic search.
Two ways of combining learning and genetic search are explored: one exploits the Baldwin effect, while the other uses a Lamarckian strategy.

Our experiments show that using a Lamarckian operator makes the algorithm find networks with a low error rate, and the smallest size, while using the Baldwin effect obtains MLPs with the smallest error rate, and a larger size, taking longer to reach a solution. 

Both approaches obtain a lower average error than other BP-based algorithms like RPROP, other evolutionary methods and fuzzy logic based methods.
\end{abstract}

\bigskip
\begin{keywords}
Evolutionary Algorithms, Generalization, Learning, Neural Networks, Optimization, Baldwin Effect, Lamarckian Search
\end{keywords}

%%%%%%%%%%%%%%%%%%%%%%%%%%%%%%%% Introduction %%%%%%%%%%%%%%%%%%%%%%%%%%%%%%%%%

\section{Introduction and State of the Art}

Hybrid algorithms often implement non-Darwinian ideas, e.g. Lamarckian evolution or the Baldwin effect, where learning influences evolution.

Lamarck's theory states that the characteristics an individual acquires during its life are passed to the offspring \cite{Lamarck09}. Thus, the following generation will inherit any acquired or learned characteristic, this mechanism would be responsible for the evolution of species. According to this approach, learning has a great influence on evolution, since all the characteristics learned are passed on to the following generation.

Nevertheless, Baldwin \cite{Baldwin96} and Waddington \cite{Waddington42} argued that this influence is limited to the fact that the individuals with greater learning capacity will adapt better to the environment, and thus will live longer. The longevity they acquire allows them to have more offspring through time, and propagate their abilities. As the number of offspring who have acquired the ability grows, this characteristic becomes part of the genetic code.

These ideas have previously been used by numerous researchers in different approaches:
\begin{itemize}
	\item \emph{Lamarckian mechanisms in hybrid evolutionary algorithms}. Lamarckian theory is today totally discredited from the biological point of view, but it is possible to implement Lamarckian evolution in EAs, so that an individual can modify its genetic code during or after fitness evaluation (its ``lifetime''). These ideas have been used by several researchers with particular success in problems where the application of a local search operator obtains a substantial improvement (travelling salesman problem, Gorges-Schleuter \cite{Gorges97}, Merz and Freisleben \cite{Merz97}, Ross \cite{Ross99}). In general, hybrid algorithms are nowadays acknowledged as the best solution to a wide array of optimization problems.

	\item \emph{Studying the Baldwin effect in hybrid algorithms} \cite{Hinton87,Belew89,Harvey93,Ackley92,Boers95}. Some authors have studied the Baldwin effect, carrying out a local search on certain individuals to improve their fitness without modifying the genetic code of the individual. This is the strategy proposed by Hinton and Nowlan in \cite{Hinton87}, who found that learning alters the shape of the search space in which evolution operates and that the Baldwin effect allows learning organisms to evolve much faster than their nonlearning equivalents, even though the characteristics acquired by the phenotype are not communicated to the genotype. 
Ackley and Littman \cite{Ackley92} studied the Baldwin effect in an artificial life system, obtaining the result that experiments in which the individuals had learning capabilities obtained the best results. Boers et al. \cite{Boers95} describe a hybrid algorithm to evolve ANN architectures, whose effectivity is explained with the Baldwin effect, implemented not as a process of learning in the network, but changing the network architecture as part of the learning process.

	\item \emph{Comparative studies of Lamarckian mechanisms and the Baldwin effect in hybrid algorithms}. Some studies have investigated whether a strategy based on a hybrid algorithm that takes advantage of the Baldwin effect is better or worse than one implementing Lamarckian mechanisms to accelerate the search \cite{Huesken2000}. The results obtained are different, and very dependent on the problem.
Gruau and Whitley \cite{GruauWhitley93} compared Baldwinian, Lamarckian and Darwinian mechanisms implemented in a genetic algorithm that evolves ANNs, finding that the first and the second strategies are equally effective for solving their problem. Nevertheless, for another problem, the results obtained by Whitley et al. \cite{WhitleyGordon94} show that taking advantage of the Baldwin effect can find the global optimum, while a Lamarckian strategy, although faster, usually converges to a local optimum.

On the other hand, results obtained by Ku and Mak \cite{KuMak97} with a GA designed to evolve recurrent neural networks, show that the use of a Lamarckian strategy implies an improvement of the algorithm, while the Baldwin effect does not. 
In Houck et al. \cite{Houck97} several algorithms are studied, and similar conclusions drawn, as in \cite{Julstrom99}, where a comparison between the Darwinian, Baldwinian and Lamarckian mechanisms, applied to the 4-cycle problem, is made. 

\end{itemize}

G-Prop (a genetic evolution of BP trained MLP), used in this paper to tune learning parameters and to set the initial weights and hidden layer size of a MLP, searches for the optimal set of weights, the optimal topology and learning parameters, using an EA and Quick-Propagation \cite{FahlmanQP} (QP). 
In this method no ANN parameters have to be set by hand; it obviously needs to set the EA constants, but is robust enough to obtain good results under the default parameter settings (all operators applied with the same probability, 300 generations and 200 individuals in the population).

%With G-Prop we intend to make use of the abilities of both families of algorithms: the ability of EA to find a solution close to the global optimum, and the ability of the BP to tune a solution and reach the closest local minimum by means of local search from the solution found by the EA. Instead of using a pre-established topology, the population is initialized with different hidden layer sizes, with some specific operators designed to change them. Mutation, multi-point crossover, addition, elimination and substitution of hidden units, and QP training applied as operator to the individuals of the population, are used. Thus, the EA searches and optimizes the architecture (number of hidden units), the initial weight setting for that architecture and the learning rate for that net as genetic operators.

This paper carries out a study of the Baldwin effect in the G-Prop \cite{CastilloIWANN99,CastilloNeurocomputing,CastilloNPL,CastilloEC2001} method to solve pattern classification and function approximation problems. 
We compare results with those of other authors, and intend to check the results obtained by Gruau and Whitley \cite{GruauWhitley93}, i.e., that the use of learning that modifies fitness without modifying the genetic code improves the task of finding an ANN to solve the problem at hand.

We compare the results obtained taking advantage of the Baldwin effect with those obtained using a Lamarckian local search mechanism.
We will also compare them with other non-hybrid (RPROP \cite{Riedmiller93}), hybrid algorithms, and those based on fuzzy logic, to prove that both versions of G-Prop obtain better results (or at least comparable) than other methods, although one of these versions is more likely to be trapped at a local optimum due to the fact that it uses a local search genetic operator.

The remainder of this paper is structured as follows: 
Section \ref{sec:method} presents the new fitness functions designed to determine if the Baldwin effect takes place in G-Prop.
Section \ref{sec:expe} describes the experiments, 
Section \ref{sec:results} presents the results obtained, 
followed by a brief conclusion in Section \ref{sec:conclus}.

%%%%%%%%%%%%%%%% Method %%%%%%%%%%%%%%%
\section{The G-Prop Algorithm}
\label{sec:method}

In this section we will only describe the new fitness functions designed to determine if the Baldwin effect takes place in G-Prop. The complete description of the method and results on classification problems have been presented elsewhere \cite{CastilloIWANN99,CastilloNeurocomputing,CastilloNPL,CastilloEC2001}.

\medskip

In G-Prop, the \emph{Darwinian fitness function} is given by the classification / approximation ability obtained when carrying out the validation after training, and in the case of two individuals with identical ability the best is the one that has a hidden layer with fewer neurons, which implies greater speed when training and classifying and facilitates its hardware implementation. 

The classification accuracy or number of hits is obtained by dividing the number of hits among the total number of examples in the validating set. The approximation ability is obtained using the normalized mean squared error (NMSE) given by:
	\begin{eqnarray}
	NMSE = \sqrt{\frac{\sum_{i}^{N}{(s_i - o_i)^2}}{\sum_{i}^{N}{(s_i - \bar{s})^2}}}
	\label{ecu:NMSE}
	\end{eqnarray}
\noindent where $s_i$ is the real output for the example $i$, $o_i$ is the obtained output, and $\bar{s}$ is the mean of all the real outputs.
%NMSE has been used because it is an index that evaluates the approximation degree independently of scale factors.

The Lamarckian approach uses no special fitness function; instead, a local search genetic operator (QP application) has been designed to improve the individuals, saving the individual trained weights (acquired characteristics) back to the population.

On the other hand, the \emph{Baldwin effect requires} some type of learning to be applied to the individuals, and the changes (trained weights) are not codified back to the population. In order to take advantage of the Baldwin effect, the following fitness function is proposed: firstly the classification/approximation ability on the validation set of the individual before being trained is calculated. Then it is trained and its ability (after training) is calculated. 
Three criteria are used to decide which is the best individual: the best MLP is that with higher classification/approximation ability after training; if both MLPs show the same accuracy, then the best is that whose ability before training is higher (the MLP is more likely to have a high accuracy when trained); if both MLPs show the same accuracy before and after training, then the best is the smallest one.

%%%%%%%%%%%%%%%% experiments %%%%%%%%%%%%%%%
\section{Experiments}
\label{sec:expe}

The algorithm was run for a fixed number of generations. 
When training each individual of the population to obtain its fitness, a limit of epochs was established. 
We used 300 generations and 200 individuals in the population in every run, and 200 training epochs in order to avoid long simulation times and also to avoid overfitted networks, making the EA carry out the search and the training operator refine the solutions. 
In addition, the number of epochs chosen was much smaller than that necessary to train a single MLP, so that the time taken to find a suitable network to solve the problem is similar to that would be needed to train a MLP (that obtains similar results) using a method based on gradient descent.
After an exhaustive test of genetic operators, we have considered to apply them with the same priority (see \cite{CastilloIWANN99,CastilloNeurocomputing,CastilloNPL,CastilloEC2001}).
The learning operator (see \cite{CastilloNPL,CastilloEC2001}) was only used when obtaining the results of the Lamarckian approach.

\medskip

The tests used to assess the accuracy of a method must be chosen carefully, because some of them (exclusive-or problem) are not suitable for certain capacities of the BP algorithm, such as generalization \cite{FahlmanBENCHMARKS}. Our opinion, along with Prechelt \cite{Prechelt94c}, is that to test an algorithm, at least two real world problems should be used.

We have used a pattern classification problem and a function approximation problem, in order to demonstrate the capacities of the proposed method solving different kind of problems, and also to show that the Baldwin effect takes place whatever the problem at hand is.

In these experiments, the \textbf{\emph{Glass1a}} (extracted from Proben1 data sets) pattern classification problem, proposed by Prechelt \cite{Prechelt94c} and used by Gr\"{o}nroos \cite{MAG98}, is used, as well as the function approximation problem given by \textbf{\emph{equation (\ref{ecu:f2})}}.

\textbf{\emph{Glass1a}} is a problem of classification of glass types, taken from \cite{Prechelt94c}. The results of chemical analysis of glass splinters (percent content of 8 different elements) plus the refractive index are used to classify the sample as either float processed or non float processed building windows, vehicle windows, containers, tableware, or head lamps. This task is motivated by forensic needs in criminal investigation. This dataset was created based on the glass problem dataset from the UCI repository of machine learning databases (\underline{\emph{http://www.ics.uci.edu/~mlearn/MLRepository.html}}). The data set contains 214 instances. Each sample has 9 attributes plus the class attribute: refractive index, sodium, magnesium, aluminium, silicon, potassium, calcium, barium, iron, class attribute (type of glass).

Function given by \textbf{\emph{equation (\ref{ecu:f2})}} is an analytical function gathered by Cherkassky \cite{CHE96} and Sugeno \cite{SUG93}, and used by Pomares \cite{Pomares99} their research on fuzzy logic function approximation:

	\begin{eqnarray}
	f_2(x) = e^{-5x} sin(2 \pi x)
	\label{ecu:f2}
	\end{eqnarray}

\noindent where $x \in [0,1]$.

%%%%%%%%%%%%%%%% results %%%%%%%%%%%%%%%
\section{Results}
\label{sec:results}

\begin{figure}[!h]
\begin{center}
\begin{tabular}{cc}
\epsfig{file=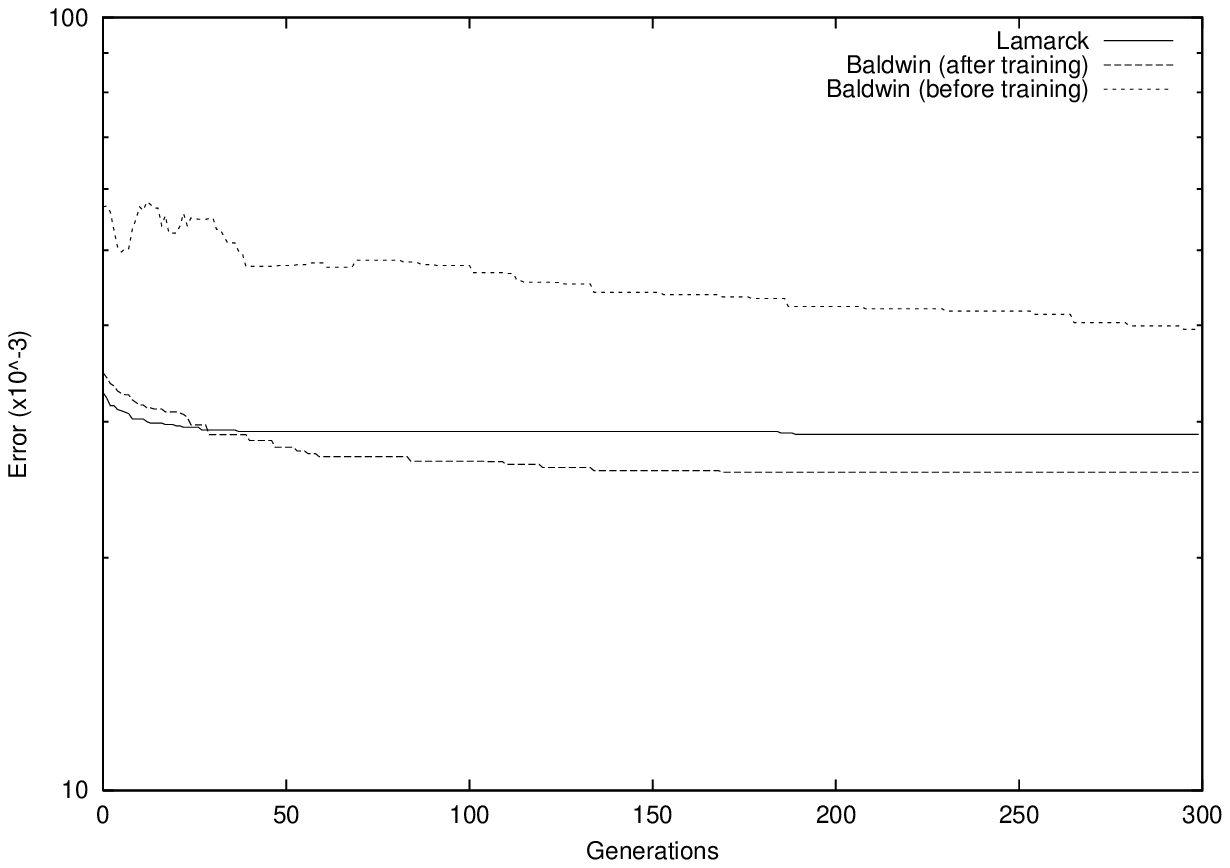,width=.5\textwidth,height=.3\textheight} \\
Error \\
\epsfig{file=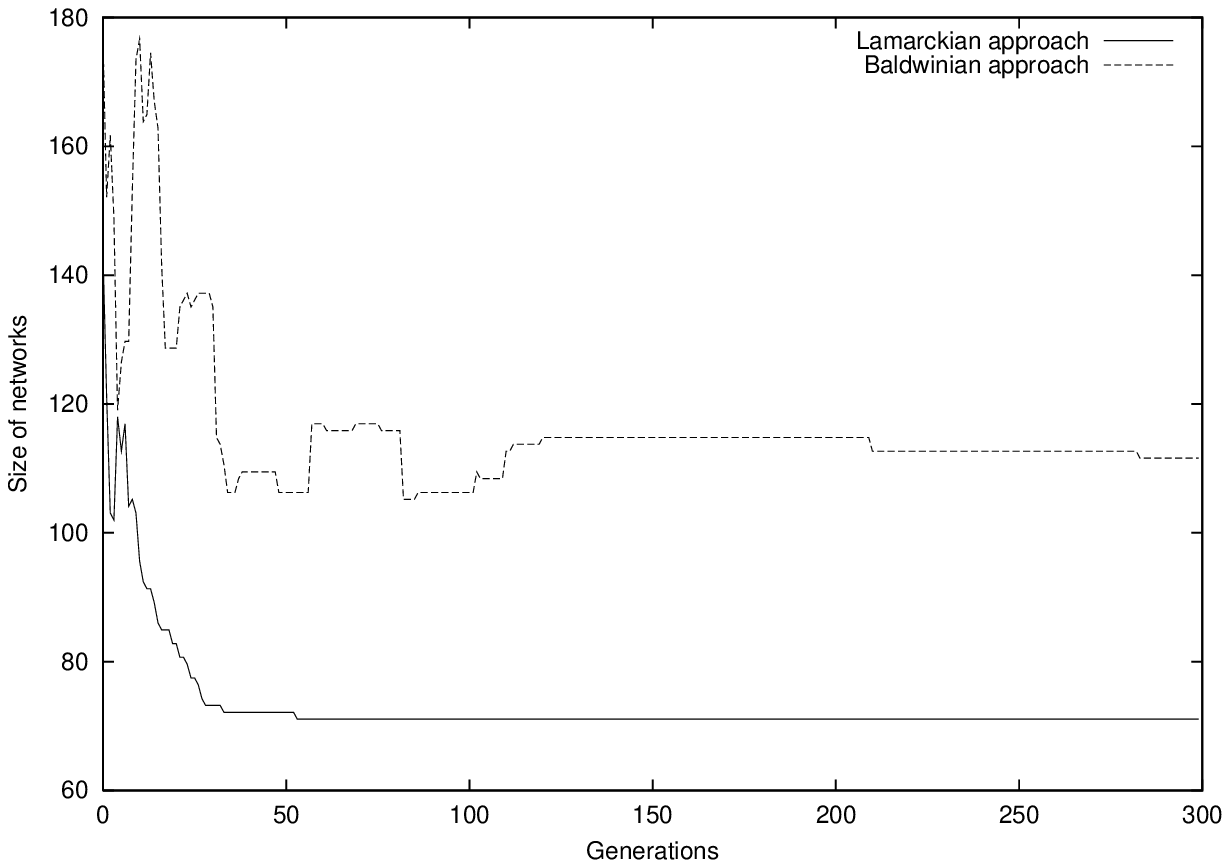,width=.5\textwidth,height=.3\textheight} \\ 
Size
\end{tabular}
\end{center}
\caption{\small{Average results over all the runs for both Lamarckian and Baldwinian approaches for the Glass problem. Average error is plotted above (on a vertical logscale) and size below.}}
\label{fig:glass}
\end{figure}

Figure \ref{fig:glass} shows average results over all the runs for both Lamarckian and Baldwinian approaches to classification and approximation problems. Plotted data correspond to the best individual in the population for each generation. The dotted line corresponds to the classification / approximation ability before training, while the dashed line corresponds to the classification / approximation ability after training, in the Baldwinian approach. The solid line corresponds to the Lamarckian approach.

Standard deviation values have not been plotted due to the fact that those values remain constant along the generations; in any case, they show that error achieved is similar.

Using Lamarckian evolution a suitable MLP is found in the early epochs of the simulation. However that MLP remains the best during the simulation (evolution stops) because of the use of an ``elitist'' algorithm, and tends to dominate the population due to its high fitness.

On the other hand, using the Baldwin effect, results can be as good as using Lamarckian evolution, although the method needs many more generations and the evolution of the population is much more progressive during the simulation. Results in size show that using the Lamarckian approach, MLPs are smaller than with the Baldwinian approach. 

\medskip

The method exhibits roughly the same behaviour on function approximation problem \textbf{$f_2$}.

\medskip

Although it is not the aim of this paper to compare the G-Prop method with those of other authors, we do so in order to prove the capacity of both versions of G-Prop to solve pattern classification and function approximation problems, and how it outperforms other methods.

Tables \ref{table:glass} and \ref{table:f2} show the average error rate, the average size of nets as the number of parameters, that is, the number of weights of the net, and the average number of generations until the best one for that run is found.

\medskip

The results for the \emph{Glass1a} pattern classification problem ($\%$ of error in test), obtained using the Lamarckian mechanism are compared with those obtained taking advantage of the Baldwin effect and those obtained by Prechelt \cite{Prechelt94c} (using {RPROP} \cite{Riedmiller93,Riedmiller94}) and Gr\"{o}nroos \cite{MAG98} (using a hybrid algorithm) in Table \ref{table:glass}.

\begin{table}[!h]
\begin{center}
\begin{tabular}{|c||c|c|c|c|c|}
\hline 
\small{Approach}     & \small{Error}          & \small{Size}          &  \small{Generations} \\
\hline
\hline
\small{Lamarckian}   &  \small{32  $\pm$ 2}   &  \small{59 $\pm$ 28}  &  \small{52 $\pm$ 54}  \\
\hline
\small{Baldwinian}   &  \small{31  $\pm$ 2}   &  \small{112 $\pm$ 62} &  \small{119 $\pm$ 55}  \\

\hline
\small{Prechelt}     &  \small{33 $\pm$ 5}        &  \small{350}      &          -           \\
\hline
\small{Gr\"{o}nroos} &  \small{32 $\pm$ 5}        &  \small{350}      &          -           \\ 

\hline
\end{tabular}
\end{center}
\caption{\small{Results for the Glass1a problem obtained with G-Prop taking advantage of the Baldwin effect and for the Lamarckian approach, as well as those obtained by Prechelt and Gr\"{o}nroos, which are included for the sake of comparison.}}
\label{table:glass}
\end{table}

It is evident that G-Prop outperforms other methods (both in classification accuracy and network size obtained): Prechelt \cite{Prechelt94c} using RPROP \cite{Riedmiller93,Riedmiller94} obtained a classification accuracy of $33 \pm 5$, and Gr\"{o}nroos \cite{MAG98} using Kitano's network obtained $32 \pm 5$; while G-Prop achieves an error of $32 \pm 2$ using the Lamarckian approach and $31 \pm 2$ taking advantage of the Baldwin effect.

In the case of the configuration that verifies if the Baldwin effect takes place in G-Prop, the classification ability obtained is greater, although the size is greater and more generations are needed to reach similar results. 

\medskip

The results for the \emph{$f_2$} function approximation problem, obtained using the Lamarckian mechanism, are compared with those obtained taking advantage of the Baldwin effect and those obtained by Pomares \cite{Pomares99} in Table \ref{table:f2}.

\begin{table}[!h]
\begin{center}
\begin{tabular}{|c||c|c|c|c|c|}
\hline 
\small{Approach}     & \small{Error}             & \small{Size}             &  \small{Generations} \\
\hline
\hline
\small{Lamarckian}   &  \small{0.09 $\pm$ 0.01}   &  \small{18 $\pm$ 8}     &  \small{34 $\pm$ 28}   \\
\hline
\small{Baldwinian}   &  \small{0.086 $\pm$ 0.004} &  \small{85 $\pm$ 27}    &  \small{97 $\pm$ 81}  \\

\hline
\small{Pomares} \cite{Pomares99} &  \small{0.125}   &   \small{6  (4 rules)}   &     -      \\

\hline
\end{tabular}
\end{center}
\caption{\small{Results for the $f_2$ problem obtained with G-Prop taking advantage of the Baldwin effect and for the Lamarckian approach, as well as those obtained by Pomares, which are included for the sake of comparison.}}
\label{table:f2}
\end{table}

The proposed method obtains better results on approximation ability ($0.09 \pm 0.01$ versus $0.125$) although the networks obtained are greater in size and number of parameters than those obtained using fuzzy controllers ($18 \pm 8$ versus $6$).

The approximation ability obtained is greater using the configuration that verifies if the Baldwin effect takes place in G-Prop, while the network sizes are slightly larger and need more generations to reach similar results.

\medskip

Each run of the proposed method takes about 4 hours on an AMD-K7(tm) 600Mhz, using the parameters described above.

The Lamarckian strategy achieves good enough results using, on average, fewer generations, although the Baldwinian strategy, using a suitable number of generations, can achieve the same or even better results.

The results obtained show that if the problem does not have many local minima and results must be obtained quickly, the best strategy is the Lamarckian. Otherwise, the Baldwinian strategy or a mixture of both is the best.

%%%%%%%%%%%%%%%% conclusions %%%%%%%%%%%%%%%
\section{Conclusions}
\label{sec:conclus}

A study of the Baldwin effect in the G-Prop method \cite{CastilloIWANN99,CastilloNeurocomputing,CastilloNPL,CastilloEC2001} (a hybrid algorithm to tune learning parameters, initial weights and hidden layer size of a MLP using an EA and QP) has been carried out.
A comparison between the results obtained taking advantage of the Baldwin effect and those obtained using a local search Lamarckian mechanism has been made.

The results obtained, agree with those presented by Whitley et al. \cite{WhitleyGordon94}, and show that the use of a Lamarckian strategy makes the method obtain good solutions faster than if the Baldwin effect is used, although it is more likely to be trapped in a local optimum than the approach that takes advantage of the Baldwin effect.
However, errors are not significatively worse.

Figures show how a Lamarckian strategy finds a suitable MLP in the early generations which remains the best during the simulation (evolution stops); with the Baldwin effect, results can be as good as those of Lamarckian evolution, although the method needs many more generations and the evolution is much more progressive.

It should be observed also that neural nets obtained using a Lamarckian strategy are smaller, which contributes to learning speed. Besides a small network is fast while training and classifying, and obtaining it in fewer generations means less time is needed to design it.

Another interesting result is that when the Lamarckian training operator is used, learning contributes more to fitness improvement at the beginning of the simulations \cite{Oliveira99}. 
This is due to to the use of an elitist algorithm, so that when it is applied to a MLP, these individuals can obtain an advantage in relation to the remaining members of the population, and will continue to be the best individuals among the population until the end of the simulation. This can be also proved using visualization techniques \cite{RomeroPPSN2000}.

%%%%%%%%%%%%%%%% acknowledgements %%%%%%%%%%%%%%%
\section*{Acknowledgements}
\label{sec:ackn}
This work has been supported in part by CICYT TIC99-0550, INTAS 97-30950, HPRN-CT-2000-00068  and IST-1999-12679.

%%%%%%%%%%%%%%%%%%%%%%%%%%%%%%%% Bibliograpyh %%%%%%%%%%%%%%%%%%%%%%%%%%%%%%%%%
\bibliographystyle{jornadas} 
\bibliography{baldwin,my_papers}
%%%%%%%%%%%%%%%%%%%%%%%%%%%%%%%%%%%%%%%%%%%%%%%%%%%%%%%%%%%%%%%%%%%%%%%%%%%%%%%

\end{document}